\documentclass[10pt,twocolumn,letterpaper]{article}

\usepackage{cvpr}              % To produce the CAMERA-READY version

\usepackage{graphicx}
\usepackage{amsmath}
\usepackage{amssymb}
\usepackage{booktabs}
\usepackage{mathrsfs}
\usepackage{ulem}
\usepackage{multirow}
\usepackage{ulem}
\usepackage{enumitem}
\usepackage{hyperref}
\usepackage{setspace}

\usepackage[capitalize]{cleveref}
\crefname{section}{Sec.}{Secs.}
\Crefname{section}{Section}{Sections}
\Crefname{table}{Table}{Tables}
\crefname{table}{Tab.}{Tabs.}
\usepackage[accsupp]{axessibility}

\usepackage{tabularx}
\usepackage{booktabs}

\def\confName{CVPR}
\def\confYear{2023}

\makeatletter
\newcommand{\printfnsymbol}[1]{%
        \textsuperscript{\@fnsymbol{#1}}%
}
\makeatother

\makeatletter
\DeclareRobustCommand\onedot{\futurelet\@let@token\@onedot}
\def\@onedot{\ifx\@let@token.\else.\null\fi\xspace}

\def\etal{et~al\onedot}

\makeatother

\begin{document}

\title{PATS: Patch Area Transportation with Subdivision for Local Feature Matching}

\author{Junjie Ni$^{1,2}$\thanks{Junjie Ni and Yijin Li contributed equally to this work.}, Yijin Li$^{1,2}$\printfnsymbol{1}, Zhaoyang Huang$^{3}$, Hongsheng Li$^{3}$, Hujun Bao$^{1,2}$,\\Zhaopeng Cui$^{1}$ and Guofeng Zhang$^{1,2}$\thanks{Guofeng Zhang is the corresponding author.} \\
$^{1}$State Key Lab of CAD\&CG, Zhejiang University ~~~~~~$^{2}$ZJU-SenseTime Joint Lab of 3D Vision \\
$^{3}$Multimedia Laboratory, The Chinese University of Hong Kong
}

\maketitle

\begin{abstract}
Local feature matching aims at establishing sparse correspondences between a pair of images.
Recently, detector-free methods present generally better performance but are not satisfactory in image pairs with large scale differences.
In this paper, we propose Patch Area Transportation with Subdivision (PATS) to tackle this issue.
Instead of building an expensive image pyramid, we start by splitting the original image pair into equal-sized patches and gradually resizing and subdividing them into smaller patches with the same scale.
However, estimating scale differences between these patches is non-trivial since the scale differences are determined by both relative camera poses and scene structures, and thus spatially varying over image pairs. Moreover, it is hard to obtain the ground truth for real scenes.
To this end, we propose patch area transportation, which enables learning scale differences in a self-supervised manner.
In contrast to bipartite graph matching, which only handles one-to-one matching, our patch area transportation can deal with many-to-many relationships.
PATS improves both matching accuracy and coverage, and shows superior performance in downstream tasks, such as relative pose estimation, visual localization, and optical flow estimation. The source code is available at
\url{https://zju3dv.github.io/pats/}.
\end{abstract}

\vspace{-0.2cm}
\section{Introduction}
\label{sec:intro}

Local feature matching between images is essential in many computer vision tasks which aim to establish correspondences between a pair of images.
In the past decades, local feature matching~\cite{orb,surf} has been widely used in a large number of applications such as structure from motion (SfM)~\cite{colmap,enft}, simultaneous localization and mapping (SLAM)~\cite{orb-slam,liu2016robust,yang2022vox}, visual localization~\cite{hloc,vsnet}, object pose estimation~\cite{rnnpose,gcasp}, etc.
The viewpoint change from the source image to the target image may lead to scale variations, which is a long-standing challenge in local feature matching.
Large variations in scale leads to two severe consequences:
Firstly, the appearance is seriously distorted, which makes learning the feature similarity more challenging and impacts the correspondence accuracy.
Secondly, there may be several pixels in the source image corresponding to pixels in a local window in the target image.
However, 
existing methods~\cite{orb,sift} only permit one potential target feature to be matched in the local window,
and the following bipartite graph matching only allows one source pixel to win the matching.
The coverage of correspondences derived from such feature matches is largely suppressed and will impact the downstream tasks.

\begin{figure}
\vspace{-0.5cm}
\begin{center}

\resizebox{0.95\linewidth}{!}{
\includegraphics[width=1.0\linewidth]{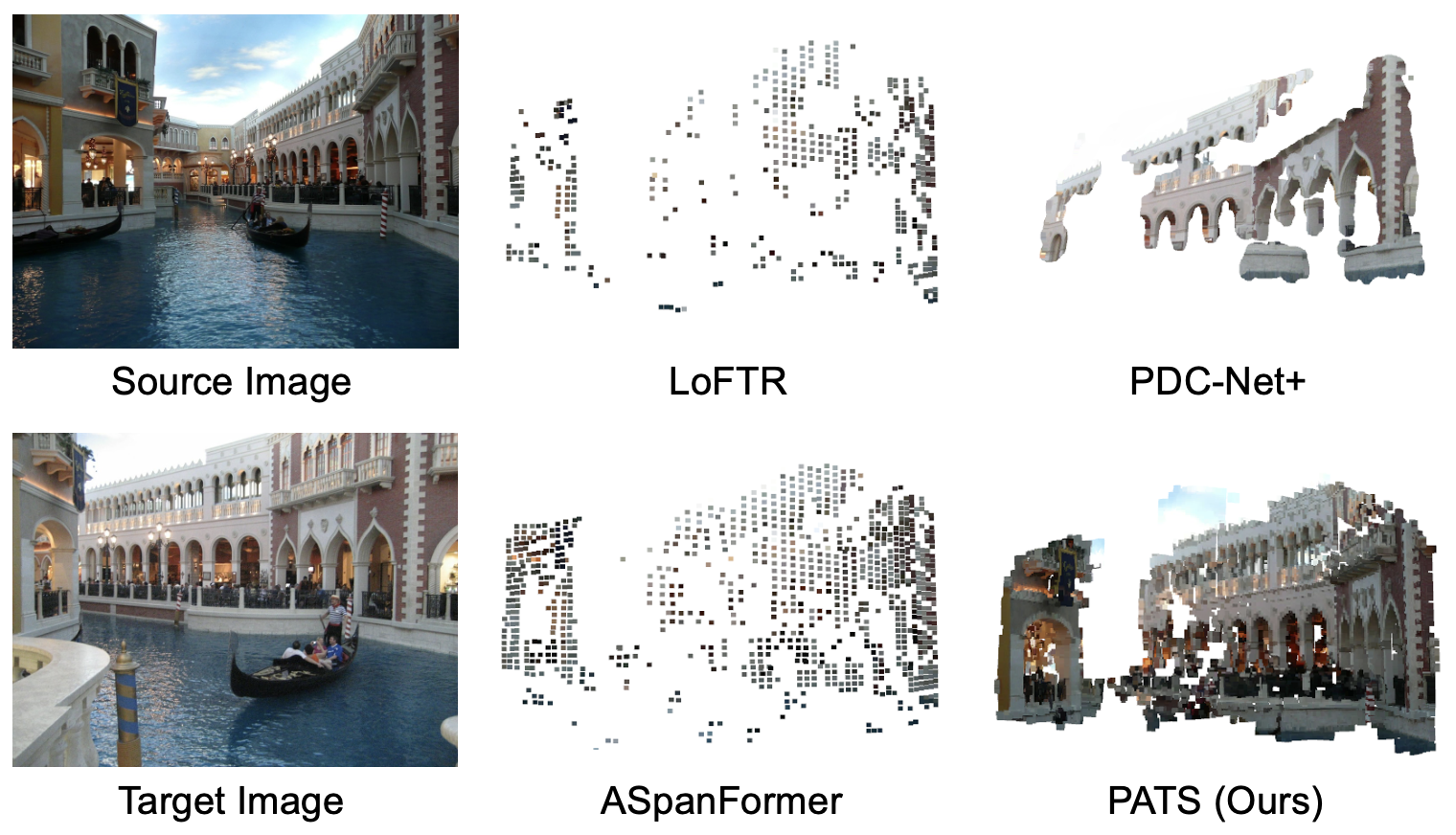}
}
\end{center}
\vspace{-0.7cm}
   \caption{\textbf{Two-view reconstruction results of LoFTR~\cite{loftr}, ASpanFormer~\cite{aspanformer}, PDC-Net+~\cite{pdcnet++} and our approach on MegaDepth dataset~\cite{megadepth}.} PATS can extract high-quality matches under severe scale variations and in indistinctive regions with repetitive patterns, which allows semi-dense two-view reconstruction by simply triangulating the matches in a image pair. In contrast, other methods either obtain fewer matches or even obtain erroneous results.}
\vspace{-0.7cm}

\label{fig:teaser}

\end{figure}

Before the deep learning era, SIFT~\cite{sift} is a milestone that tackles the scale problem by detecting local features on an image pyramid and then matching features crossing pyramid levels, called scale-space analysis.
This technique is also adopted in the inference stage of learning-based local features~\cite{r2d2}.
Recently, LoFTR abandons feature detection stage and learns to directly draw feature matches via simultaneously encoding features from both images based on the attention mechanism.
By removing the information bottleneck caused by detectors, LoFTR~\cite{loftr} produces better feature matches.
However, LoFTR does not handle the scale problem
and the scale-space analysis is infeasible in this paradigm because conducting attention intra- and inter- different scales will bring unbearable increasing computational costs.
As a result, the scale curse comes back again.

\begin{figure}
% \vspace{-0.5cm}
\begin{center}
\resizebox{0.95\linewidth}{!}{
\includegraphics[width=1.0\linewidth]{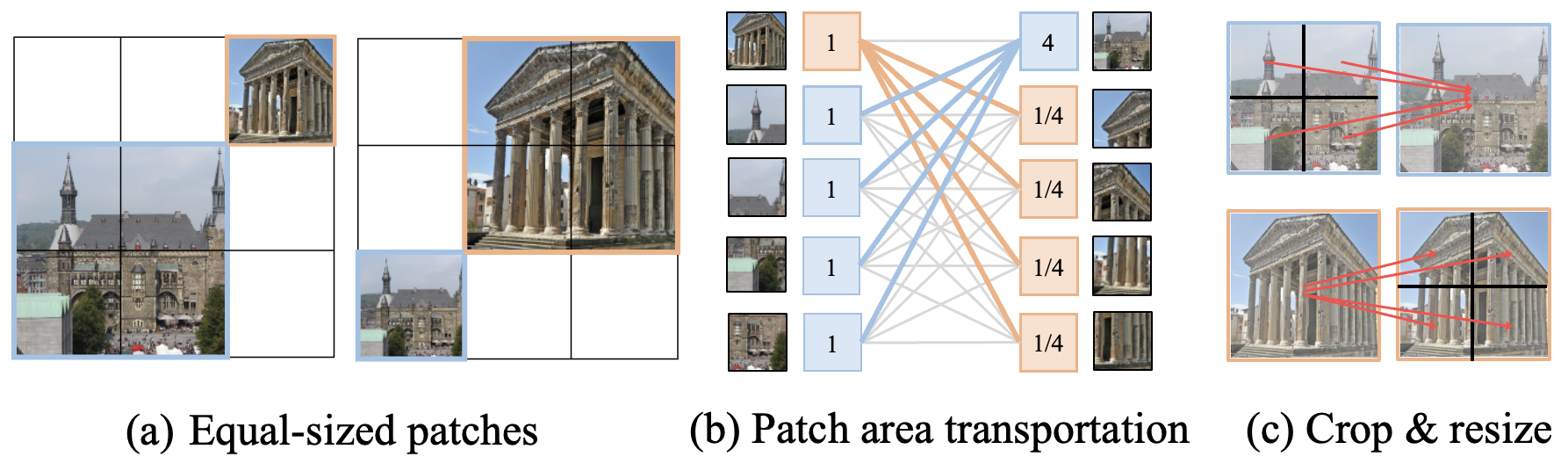}
}
\end{center}
\vspace{-0.7cm}
   \caption{\textbf{Scale Alignment with Patch Area Transportation.}
   Our approach learns to find the many-to-many relationship and scale differences through solving the patch area transportation. Then we crop the patches and resize the image content to align the scale, which remove the appearance distortion.
   }
   % In this case, the proposed patch area transportation learn to find the one-to-four and four-to-one patch relationships in a self-supervised way, which }
   % By a) splitting images into patches, b) regressing target patches' area, and finding the one-to-four and four-to-one patch relationships with the patch area transportation, we can c) crop the source patches with their corresponding target regions. Then, the cropped target region is resized to align the source patches' size. The appearance distortion is removed, and all the matches are reserved, which eases feature matching and increases coverage.}
\vspace{-0.7cm}
\label{fig:intro}

\end{figure}

In this paper, we propose \textbf{P}atch \textbf{A}rea \textbf{T}ransportation with \textbf{S}ubdivision (PATS) to tackle the scale problem in a detector-free manner.
The appearance distortion can be alleviated if the image contents are aligned according to their scale differences before feature extraction.
As shown in Fig.~\ref{fig:intro}, if the target image is simply obtained by magnifying the source image twice, a siamese feature encoder will produce features with large discrepancies at corresponding locations.
The discrepancies are corrected if we estimate the scale difference and resize the target image to half before feature extraction.
Considering that scale differences are spatially varying, 
we split the source image into equal-sized patches and then align the scale patch-wisely.
Specifically, we identify a corresponding rectangular region in the target image for each source patch and then resize the image content in the region to align the scale.
By also splitting the target image into patches, the rectangular regions can be represented with patches bounded by boxes.
Based on this representation, one source patch corresponds to multiple target patches. 
Moreover, the bounding box may be overlapped, indicating that one target patch may also correspond to multiple source patches.
Here comes the question: how can we find many-to-many patch matches instead of one-to-one~\cite{superglue,loftr,aspanformer}?

We observe that finding target patches for a source patch can be regarded as transporting the source patch to the target bounding box, where each target patch inside the box occupies a portion of the content.
In other words, the area proportion that the target patches occupying the source patch should be summed to 1.
Motivated by this observation, we propose to predict the target patches' area and formulate patch matching as a patch area transportation problem that transports areas of source patches to target patches with visual similarity restrictions.
Solving this problem with Sinkhorn~\cite{sinkhorn}, a differential optimal transport solver, also encourages our neural network to better capture complex visual priors. 
Once the patch matching is finished, the corresponding bounding boxes can be easily determined.
According to the patch area transportation with patch subdivision from coarse to fine,
PATS significantly alleviates appearance distortion, which largely eases the difficulty of feature learning to measure visual similarity.
Moreover, source patches being allowed to match overlapped target patches naturally avoid the coverage reduction problem.
After resizing the target regions according to estimated scale differences, we subdivide the corresponding source patch and target region to obtain finer correspondences, dubbed as scale-adaptive patch subdivision.
Fig.~\ref{fig:teaser} shows qualitative results of our approach.

Our contributions in this work can be summarized as three folds:
1) We propose patch area transportation to handle the many-to-many patch-matching challenge and grants the ability that learning scale differences in a self-supervised manner to the neural network.
2) We propose a scale-adaptive patch subdivision to effectively refine the correspondence quality from coarse to fine.
3) Our patch area transportation with subdivision (PATS) achieves state-of-the-art performance and presents strong robustness against scale variations.

\section{Related works}

\noindent \textbf{Detector-Free Feature Matching.}
Given two images to be matched, classic methods~\cite{sift,orb,surf} usually adopt three-phase pipeline: feature detection~\cite{superpoint}, feature description~\cite{d2net,r2d2}, and feature matching~\cite{superglue, goodcorr}.
The phase of feature detection reduces the search space of matching but also introduces an information bottleneck.
When the feature detector fails to extract feature points, we can not find good correspondence even with perfect descriptors and matching strategy.
Consequently, an alternative approach that concentrates on producing correspondences directly from unprocessed images, known as a detector-free framework, has emerged.
Earlier works~\cite{drcnet} in detector-free matching usually rely on cost volume~\cite{pwcnet} to enumerate all the possible matches.
Recently, Sun~\etal~\cite{loftr} propose encoding features from both images based on the Transformer~\cite{transformer,deltar,flowformer,shi2023flowformer++}, which better model long-range dependencies and achieve satisfying performance.
Another concurrent work~\cite{cotr} show that this kind of framework bridge the task of local feature matching and optical flow estimation~\cite{videoflow, neuralmarker,blinkflow}.
After that, many variants have been proposed~\cite{quadtree,matchformer,aspanformer}.
However, these works rarely focus on the scale difference between the image pair.
To fill the gap, we propose a new detector-free framework that efficiently and effectively tackles the scale problem.
We show that by removing the one-to-one matching constraint and alleviating the appearance distortion, our methods obtain a significantly larger number of matches which is also more accurate.

\noindent \textbf{Scale-Invariant Feature Matching.}
These methods attempt to address the challenges posed by the potential scale differences between the image pair.
To this end, traditional methods~\cite{sift,surf} usually detect local features on an image pyramid and then matching features crossing pyramid levels, which is called scale-space analysis\cite{scalespace}.
The technique is also adopted in the inference stage of learning-based methods~\cite{hddnet,aslfeat,r2d2}.
Recent methods tend to mitigate scale by directly predicting scale difference~\cite{scalenet}, overlap area~\cite{overlap_est}, warping deformation~\cite{viewpoint_invariant,rotation_invariant}, or iteratively estimating correspondence and computing co-visible area~\cite{cotr}.
Compared to these methods, we propose patch area transportation which simultaneously infers the scale difference and correspondence.
The structure of the area transportation enables the network to learn scale differences in a self-supervised manner and also encourages the network to better capture complex visual priors.

\noindent \textbf{Optimal Transport in Vision-Related Tasks.}
Optimal transport has been widely used in various computer vision tasks such as object detection~\cite{detr,ota}, semantic segmentation~\cite{transport_segmentation}, domain adaptation~\cite{transport_domain_ada}, shape matching~\cite{transport_shape_matching}, visual tracking~\cite{transport_visual_tracking}, semantic correspondence~\cite{transport_semantic_corres} and so on.
Most of these works seek to find the best one-to-one assignment with the lowest total cost, which is equal to bipartite matching.
The problem is a special case of transportation problem which allows many-to-many relationships.
In this paper, we show that introducing the transportation modeling to the feature matching can be of great benefit, and we hope it can further inspire research in other areas.

\begin{figure*}
\vspace{-0.6cm}
\begin{center}
\includegraphics[width=0.90\textwidth]{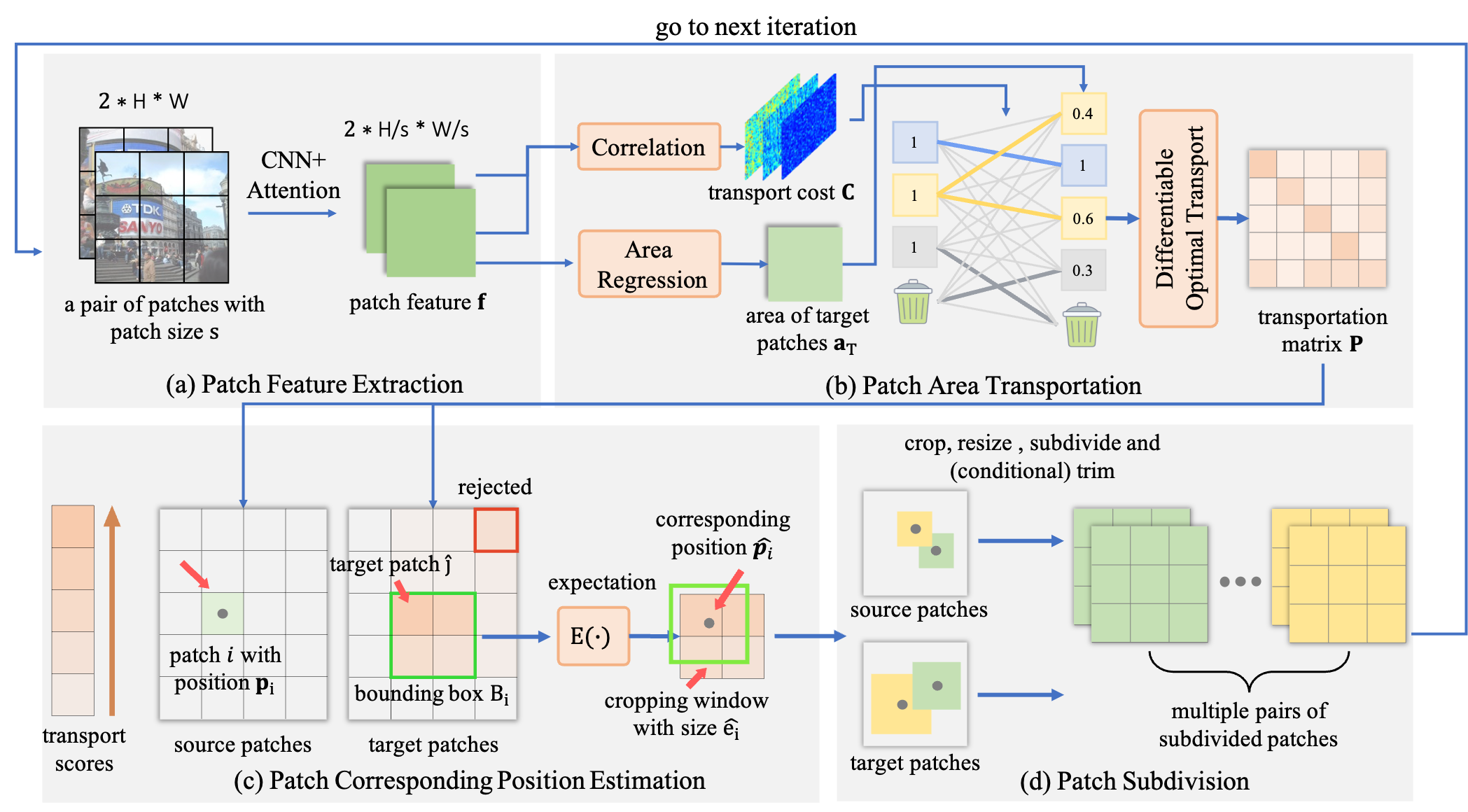}
\end{center}
\vspace{-0.6cm}
   \caption{\textbf{Overview of PATS.} We a) extract features for patches. Then, we b) formulate the patch area transportation by setting source patches' area $\mathbf a_S$ as $\mathbf 1_N$, regressing target patches' area $\mathbf a_T$, and bound the transportation via visual similarities $\mathbf C$.
   The feature descriptors $\mathbf f$ that produce $\mathbf C$ and the area regression $\mathbf a_T$ are learned by solving this problem differentially.
   The solution of this problem $\mathbf P$ also reveals many-to-many patch relationships.
   Based on $\mathbf P$, we c) find corresponding regions, represented by target patches inside a bounding box $B_i$, for each source patch. The exact patch corresponding position $\hat{\mathbf p}_i$ is the position expectation over $B_i$.
   After cropping and resizing image contents according to the obtained window sizes, which align the contents to the same scale, we d) subdivide the cropped contents to smaller patches and enter the next iteration.
   }
\vspace{-0.5cm}
\label{fig:pipeline}
\end{figure*}

\section{PATS}

As presented in Fig.~\ref{fig:pipeline}, given a pair of images, we divide them into equal-sized patches and seek corresponding patches in the target image for each patch in the source image.
One source patch transported to the target image may be stretched and cover multiple target patches,
so we regress the areas of target patches and find many-to-many patch matches via the proposed patch area transportation algorithm.
Once the patches are matched, we prune unreliable matches, crop and resize the target patches to align the source patches' scale, which alleviates the appearance distortion derived from scale variations.
Then, we conduct the patch subdivision to enter the finer level for correspondence refinement.
Finally, PATS obtains accurate correspondences according to the matched fine-grained patches.
We introduce how to extract features from patches in Sec.~\ref{sec:feature_extraction}, the patch transportation in Sec.~\ref{sec:patch-transportation}, the hierarchical patch subdivision in Sec.~\ref{sec:patch_subdivision}, and finally, explain the supervision in Sec.~\ref{sec:supervision}.

\subsection{Patch Feature Extraction}
\label{sec:feature_extraction}
Given the image pair $\mathbf I_S, \mathbf I_T \in \mathbb R^{H\times W\times 3}$, we divide them into square patches of equal size $s$. We use $\mathcal{S}:= \left\{1, ..., i, ... , N\right\}$ and $\mathcal{T}:= \left\{1, ... , j, ..., M\right\}$ to index the source patches and the target patches, respectively.
For each patch, we associate it with a 2D position $\mathbf p \in \mathbb R^2$, a descriptor $\mathbf {f} \in \mathbb R^d$, and an area $a \in \mathbb R$.
$\mathbf p$ is the image coordinate of each patch's center, and the descriptors $\mathbf {f}$ is obtained through the feature extraction module.
$a$ reflects the area of the patch when scaled to the source image.
Certainly, each source patch has the unit area ($a_i=1$).
The areas of target patches $a_j$ are regressed by our neural network.
A straightforward method to supervise $a_j$ is collecting the ground truth beforehand, but they are unavailable.
Instead, we learn to predict $a_j$ by the differential patch transportation, which will be introduced in section~\ref{sec:patch-transportation}.

Inspired by LoFTR, we first encode an image pair with a siamese convolutional neural network~\cite{siamese}, where the size of the feature map achieves $H/s\times W/s\times d$, and then encode both image features with stacked self-attention and cross-attention layers.
This step produces the descriptor $\mathbf f_i \in \mathbb R^{N\times d}$, $\mathbf f_j \in \mathbb R^{M\times d}$ for the source patches and target patches.
We have $N=M=(H/s\times W/s)$ at the beginning while $N$ and $M$ gradually increase during the following coarse-to-fine subdivision.
For all target patches, we append a small CNN to the descriptors to regress their areas $a_j$.
With such triplet features $(\mathbf p,a, \mathbf{f})$ for each patch, we propose a patch transportation method to match the patches.

\subsection{Patch Matching with Area Transportation}
\label{sec:patch-transportation}

There are non-negligible scale differences between the image pairs to be matched, which makes one source patch may correspond to multiple target patches, and vice versa.
Ignoring the scale and forcing one-to-one bipartite graph matching reduces the match number and impacts the matching accuracy.
As shown in Fig.~\ref{fig:pipeline}, when we transport a source patch to the target image, its area may be dispersed into multiple target patches, which are all corresponding patches.
To this end, we propose to formulate the patch matching as an optimal transportation problem, i.e., transporting source patches' areas to target patches and minimizing the transportation cost defined as patch similarities.

\noindent\textbf{Patch Area Transportation}. Given all patches' areas $a$ and descriptors $\mathbf{f}$, we define the cost between patches as the correlation of their descriptors:
\begin{equation}
  C_{i,j} = - \left<\mathbf{f}_i, \mathbf{f}_j\right>, \forall (i ,j) \in \mathcal{S} \times \mathcal{T},
  \label{eq:cost_compute}
\end{equation}
where $\left<\ . \ , \ .\ \right>$ is the inner product.
A part of the source patch transported to the target image may be out of border or occluded, which is invisible.
To handle these cases, we define the patch transportation as partial transportation.
Formally, we seek to figure out the transportation matrix $\mathbf P$ to minimize the cost function such that transporting the source patches' area to the target patches:
\begin{equation}
\begin{aligned} 
       \mathbf P 
   \mathbf 1_{M} \preceq
  \mathbf a_{S},\  
       \mathbf P^T \mathbf 1_{N} \preceq
  \mathbf a_{T} 
\end{aligned}
\end{equation}
where $\mathbf a_S \in \mathbb R^N, \mathbf a_T \in \mathbb R^M$ denote the area of the source patches and the target patches. As all source patches' area is 1, we have $\mathbf a_S=\mathbf 1_{N}$.
Following ~\cite{superglue,loftr}, we add a dustbin to the patches collection in both the source image and the target image to handle the partial transportation. 
The transportation cost from a source patch $i$ to a target patch $j$ is defined as $P_{i,j}\times C_{i,j}$.
The optimal transportation problem is computed efficiently with the Sinkhorn~\cite{sinkhorn} algorithm, which is differentiable. 
The transportation solution that minimizes the total transportation cost maximizes the visual similarities weighted by the transport areas.

With the restrictions derived from differential patch area transportation, our neural network simultaneously learns to infer complex area relationships and patch descriptors.
For example, transporting multiple source patches to the same target patch that is visually similar urges the area of target patches to be large such that it can accommodate these source patches. otherwise, some source patches leak to other target patches or the dustbin with a potentially higher transport cost.
Our patch transportation formulation that minimizes the transport cost, the negative of patch similarity, weighted by the transport mass, the area, also encourages our network to properly predict the feature descriptors and areas to accomplish this goal.
Compared with the bipartite graph matching that directly maximizes feature similarities, our patch transportation  introduces a better inductive bias and provides guidance for the patch subdivision.

\noindent\textbf{Patch Corresponding Position Estimation}. 
The next step is to calculate the precise positions of corresponding patches in the target image for all source patches.
After patch area transportation, we obtained the transportation matrix $\mathbf P$ that represents the relationship between each source patch and each target patch.
For a source patch $i$, we first find the target patch $\hat{j}$ that occupies the largest area of $i$: 
\begin{equation}
\hat{j}=\mathop{\arg\max}_{1\leq j\leq M}\mathbf P_{i,j}.
\end{equation}
A target patch $j'$ is feasible if the transportation area is larger than a threshold $\mathbf P_{i,j'}\geq \epsilon$.
We expand $\hat{j}$ to collect more feasible target patches $j'$ with a 4-connected flood fill algorithm.
The collection of $j'$ presents irregular shapes, so we collect target patches inside its axis-aligned bounding box as the final corresponding target patches $B_i$.
The corresponding position $\hat{\mathbf p}_i$ is the expectation over $B_i$:
\begin{equation}
\begin{aligned}
 &   w_{i,j}= \sqrt{\frac{P_{i, j}}{a_j}}, w_i = \sum_{j \in B_i}w_{i,j} \\
&  \hat{\mathbf p_i} = \sum_{j \in B_i} \frac{w_{i,j}}{w_i} \cdot \mathbf p_j,\quad \forall i \in \mathcal{S}. \\
\end{aligned}
\label{eq:interpolate}
\end{equation}
Considering the area quadratically increases with the length of side, we use the root of areas as weights. ($\mathbf p_i, \hat{\mathbf{p}}_i$) constitutes a match and will be refined during patch subdivisions.

\subsection{Patch Subdivision}
\label{sec:patch_subdivision}
Generally, the error of the target position is proportional to the size of the patch, but the smaller size of patches indicates more patches, which leads to unacceptable computations and memory in transformers.
To this end, we start with a large patch size and adopt the coarse-to-fine mechanism.
With $\{(\mathbf{p}_i, \hat{\mathbf{p}}_i, B_i)|i \in \mathcal S\}$ obtained by the patch transportation at level $l$, we crop a local window around each source and target position, resize the target window to the size of the source window, and subdivide the windows into smaller patches for the next level.
To ensure the cropped source-target window pairs share the same image content, we need to carefully compute their window size.
The window size for a source patch is naturally defined by its patch size, but how to determine the window size for its corresponding target position is not easy.
Fortunately, we can compute the scaling factor $\gamma_i$ from the source position to the target position via the obtained areas.
With the scaling factor, we can determine the target window size from the source window size.
For convenience, we omit the superscript $l$ of the variables at the $l$ level.

\begin{figure}
\vspace{-0.3cm}
\begin{center}
\resizebox{1.0\linewidth}{!}{
\includegraphics[width=0.5\textwidth]{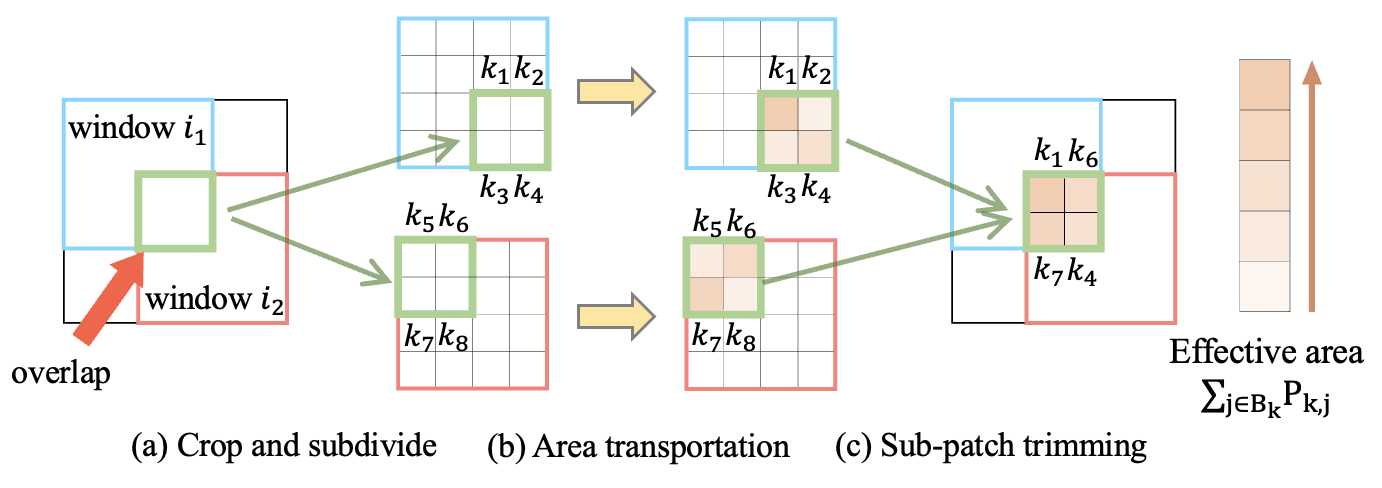}
}
\end{center}
\vspace{-0.7cm}
\caption{\textbf{Sub-patches Trimming.} 
a) The windows of neighboring source patches are partially overlapped due to the expansion. b) After subdivision, the sub-patches at the overlapped locations are redundant. c) We reserve the sub-patches that send the largest effective area $\sum_{j\in B_k} P_{k,j}$ as patches for the next level.}
\label{fig:merging}
\vspace{-0.5cm}
\end{figure}

\noindent\textbf{Patch Cropping and Subdivision.}
Given the matched patch $(\mathbf p_i, \hat{\mathbf p}_i)$, 
we estimate their scaling factor $\gamma_i$ and crop different-sized local windows around them with respect to the scale differences.
Specifically, the source window size $e$ is fixed, determined by the patch size $s$ at this level.
To determine the target window size of a source patch, we first compute the area expectation of the target position:
\begin{equation}
  \label{eq:compute_scale_weights}
\begin{split}
    \hat{a}_i = \sum_{j \in B_i} \frac{P_{i,j}}{\sum_{j \in B_i} P_{i, j}} a_j.
\end{split}
\end{equation}
Note that $j\in B_i$ are the corresponding target patches we find for the source patch $i$, so $\sum_{j \in B_i} P_{i, j}$ indicates the area that we successfully transport to the target image.
$\hat{a}_i$ indicates the area of the target window received from the source patch. 
We thus are able to compute the scaling factor by: $\gamma_i = {\sqrt{a_i}}/{\sqrt{\hat{a_i}}}$ and then  compute the  target window size $\hat{e}_i= \gamma_i e$ with high confidence that the target window covers what the source window sees.
A choice for the source window size $e$ is to let it equal the patch size $s$, which means directly cropping the source patch.
However, potential errors during patch matching and the regular shape of patches may lead to missing the matching near the window boundary.
To this end, we choose to crop a larger window with the size $e=ns$ with an expansion factor $n$.
The expansion operation increases the matching recall with the cost of bringing patch overlapping, which needs to be trimmed in the following step.
After that, we resize the cropped target window to the source window size $e$, which narrows the scale differences and alleviates the appearance distortion.
As shown in the fourth part in the Fig.~\ref{fig:pipeline}, the source windows and resized target windows are finally subdivided into sub-patches with a smaller patch size $s^{l+1}$, so that each window produce $K\times K$ sub-patches ($K=e/s^{l+1}$).

\noindent \textbf{Sub-patch Trimming with Area Transportation}
We perform sub-patch trimming as shown in Fig.~\ref{fig:merging}.
Note that the total number of sub-patches ($N \times K \times K$) in the source image is larger than directly dividing the image by the patch size $s^{l+1}$ ($H/s^{l+1}\times W/s^{l+1}$) due to overlapping brought by the expansion operation.
As we find correspondence for each source patch, the overlapped source sub-patches cause ambiguous correspondence at the next level.
We thus need to trim the sub-patches $k\in \{1, ..., N\times K\times K\}$ to obtain non-repeated patches for the next level $i^{l+1}\in \{1, ..., H/s^{l+1}\times W/s^{l+1}\}$.
Here comes the question: suppose $r_{i^{l+1}}$ contains the sub-patches that are repeated on the patch for the next level $i^{l+1}$, which sub-patch $k\in r_{i^{l+1}}$ worth keeping?
Recall that $\sum_{j \in B_i} P_{i, j}$ indicates the area that we successfully transport from the source patch $i$ to the target image.
Intuitively, the area of a patch transported to the target image is small indicates the patch loses its target with a high probability, so we can regard the total transport area as confidence and keep the sub-patch owning the largest transported area.
Specifically, inside each window pair, we extract the features for sub-patches and apply the patch area transportation to compute the transportation matrix as introduced in Sec.~\ref{sec:feature_extraction} and Sec.~\ref{sec:patch-transportation}, so that we obtained corresponding $\mathbf f_k^{l+1}, B_k^{l+1}$ for source sub-patches and $\mathbf P^{l+1}$.
Here, we slightly abuse $j\in {1,...,N\times K\times K}$ to denote the index of sub-patches in target windows.
Suppose the sub-patches repeated in $i^{l+1}$ are collected in $r_{i^{l+1}}$, we keep the sub-patch that has the maximal area:
\begin{equation}
\hat{k}=\mathop{\arg\max}_{k \in r_{i^{l+1}}}\mathbf \sum_{j\in B_k} P_{k,j}.
\end{equation}

After sub-patch trimming, we get smaller patches, and each source patch only focuses on a small group of target patches. We can now enter the next level to refine the correspondence.

\subsection{Supervision}
\label{sec:supervision}

We collect the ground truth correspondences for supervision following LoFTR~\cite{loftr}.
Given a pair of images with their depth map and camera poses, we warp a pixel at $\mathbf{p}_i$ to the target image via the camera pose and its depth as the ground truth correspondence $\tilde{\mathbf p}_i$.
We supervise our neural network with three losses:
an outlier patch loss $L_o$, an inlier patch loss $L_i$, and a patch concentration loss $L_c$.
The estimated position $\hat{\mathbf{p}}_i$ is the expectation of patches inside the bounding box.
However, the ground truth position $\tilde{\mathbf p}_i$
maybe outside the bounding box at the beginning.
Directly supervising $\hat{\mathbf{p}}_i$ with $\tilde{\mathbf p}_i$ provides meaningless gradients, and the neural network can not achieve convergence.
We thus divide our source patches into outlier patches $\mathcal{M}_o$ and inlier patches $\mathcal{M}_i$ according to the distance error: 
\begin{equation}
\begin{aligned}
&\mathcal{M}_o=\{(i,j)|\ ||\hat{\mathbf{p}}_i-\tilde{\mathbf p}_i||^2 > \theta \}, \\ &\mathcal{M}_i=\{(i,j)|\ ||\hat{\mathbf{p}}_i-\tilde{\mathbf p}_i||^2 \leq \theta \}.
\end{aligned}
\end{equation}
Here, the index j indicates the target patch where the ground truth correspondence $\tilde{\mathbf p}_i$ locate inside.
We apply the outlier patch loss for $i \in \mathcal{M}_o$ and the inlier patch loss with the patch concentration loss for $i \in \mathcal{M}_i$.

\noindent\textbf{Outlier Patch Loss.} For the patches belonging to the outlier set, we directly encourage the source patch to transport its area to the target patch by minimizing the negative log of the corresponding transportation mass:
\begin{equation}
  L_o = - \frac{1}{|\mathcal{M}_o|}\sum_{(i, j) \in \mathcal{M}_o} log P_{i,j}.
  \label{eq:classifying loss}
\end{equation}
This outlier patch loss effectively pulls the estimated target location into the  corresponding target patch.

\noindent\textbf{Inlier Patch Loss.}
The outlier patch loss can only achieve patch-wise accuracy. For inlier patches, we directly minimize the distance for fine-grained correspondence accuracy:
\begin{equation}
  L_i = \frac{1}{|\mathcal{M}_i|}\sum_{(i, j) \in \mathcal{M}_i} ||\hat{\mathbf{p}}_j - \tilde{\mathbf{p}}_j||^2
  \label{eq:position_loss}
\end{equation}

\noindent\textbf{Patch Concentration Loss.} 
One source patch transported to the target image is expected to be clustered together.
In our PATS, the cluster is the target patches we collected in the bounding box, so the area transported from the source patch to the target image is expected to concentrate in the bounding box.
Therefore, for the target patches outside the bounding box $j'\notin B_i$, we suppress the area they received:
\begin{equation}
  L_c =  \frac{1}{|\mathcal{M}_i|}\sum_{(i, j) \in \mathcal{M}_i, j' \notin B_i}P_{i, j'}
  \label{eq:block loss}
\end{equation}

We use these three losses to supervise our model: $L = L_o + L_i + L_c.$
% \begin{equation}
% \begin{split}
%   L &= L_o + L_i + L_c.
%   \label{eq:all_losses}
% \end{split}
% \end{equation}

\begin{table}[!t]
% \vspace{-0.6cm}

\tabcolsep 4pt
\footnotesize
% \normalsize

\centering
\begin{tabular}{ccccc}
\toprule
\multirow{2}{*}{Category} & \multirow{2}{*}{Method} & \multicolumn{3}{c}{Pose estimation AUC}       \\ \cline{3-5} 
      & & @5\textdegree            & @10\textdegree           & @20\textdegree          \\ \hline
\multirow{2}{*}{Detector-based}
& \multicolumn{1}{l|}{RootSIFT~\cite{rootsift}+SGMNet~\cite{sgmnet}}              & 35.5          & 55.2          & 71.9          \\
&\multicolumn{1}{l|}{SP~\cite{superpoint}+SuperGlue~\cite{superglue}}              & 38.7          & 59.1          & 75.8          \\ \hline
\multirow{5}{*}{Detector-free}
& \multicolumn{1}{l|}{DRC-Net~\cite{drcnet}}                                        & 29.5          & 50.1          & 66.8          \\
&\multicolumn{1}{l|}{PDC-Net+~\cite{pdcnet++}}                                     & 39.1          & 60.1          & 76.5          \\
&\multicolumn{1}{l|}{LoFTR~\cite{loftr}}                                           & 42.4          & 62.5          & 77.3          \\
&\multicolumn{1}{l|}{ASpanFormer~\cite{aspanformer}}                                & \underline{44.5}    & \underline {63.8}    & \underline{78.4}    \\
&\multicolumn{1}{l|}{Ours}                                                         & \textbf{47.0} & \textbf{65.3} & \textbf{79.2} \\
\bottomrule
\end{tabular}
\vspace{-0.3cm}
\caption{Evaluation on YFCC100M~\cite{yfcc100m} for outdoor pose estimation.}
\vspace{-0.3cm}
\label{tab:yfcc}

\end{table}

\subsection{Implementation Details}
We conduct the patch transportation and the subdivision alternatively for $L=3$ times, with patch sizes of 32, 8, and 2 at each hierarchy, respectively.
We train the network progressively by adding and training hierarchy one by one while freezing the weights of all previously trained hierarchies.
The network is trained using AdamW~\cite{adamW} with an initial learning rate of 1e-4 and batch sizes of 128, 48, and 12 were set for each hierarchy.
Each hierarchy converges after 30 hours of training with 3-4 NVIDIA RTX 3090 GPUs.
We set the expansion factor $n$ to 3 and 2 for the cropping in the first and second hierarchy.
The threshold $\epsilon$ is set to 1e-5.
We set $\theta$ to be equal to the patch size at each hierarchy. 
% More details are provided in the supplementary materials.

\section{Experiments}

\begin{table}[!t]
% \vspace{-0.6cm}

\tabcolsep 5pt
\footnotesize
% \normalsize

\centering
\begin{tabular}{clccc}
% \hline
\toprule
\multirow{2}{*}{Category} & \multicolumn{1}{c}{\multirow{2}{*}{Method}} & \multicolumn{3}{c}{Pose estimation AUC}       \\ \cline{3-5} 
      & & @5\textdegree            & @10\textdegree           & @20\textdegree          \\ \hline
\multirow{3}{*}{Detector-based}
& SP~\cite{superpoint}+OANet~\cite{oanet}                    & 11.8         & 26.9         & 43.9         \\
& SP~\cite{superpoint}+SGMNet~\cite{sgmnet}                  & 15.4         & 32.1         & 48.3         \\
& SP~\cite{superpoint}+SuperGlue~\cite{superglue}            & 16.2         & 33.8         & 51.8         \\ \hline
\multirow{6}{*}{Detector-free} 
&DRC-Net~\cite{drcnet}                                      & 7.7           & 17.9          & 30.5          \\
&LoFTR~\cite{loftr}                                         & 22.1         & 40.8          & 57.6         \\
&MatchFormer~\cite{matchformer}                             & 24.3          & 43.9          & 61.4          \\
&QuadTree~\cite{quadtree}                                   & 24.9          & 44.7          & 61.8          \\
&ASpanFormer~\cite{aspanformer}                             & \underline {25.6}    & \underline {46.0}      & \underline {63.3}    \\ 
&Ours                                                       & \textbf{26.0} & \textbf{46.9} & \textbf{64.3} \\
\bottomrule
\end{tabular}
\vspace{-0.3cm}
\caption{Evaluation on ScanNet~\cite{scannet} for indoor pose estimation.}
\label{tab:scannet}

\vspace{-0.3cm}

\end{table}

\begin{table}[!t]
% \vspace{-0.2cm}
\footnotesize
\centering
\begin{tabular}{lccccc}
\toprule
\multicolumn{1}{l|}{Method}      & 320                            & 480                            & 640                            & 1024                           & 1600                           \\ \hline
\multicolumn{6}{c}{Pose estimation (AUC @5\textdegree)}                                                                                                                                                            \\ \hline
\multicolumn{1}{l|}{SP\cite{superpoint}+SuperGlue~\cite{superglue}}   & 25.5                           & 33.5                           & 38.7                           & 41.4                           & 43.0                           \\
\multicolumn{1}{l|}{LoFTR~\cite{loftr}}       & 35.9                           & 51.6                           & 54.2                           & 52.9                           & 22.2                           \\
\multicolumn{1}{l|}{ASpanFormer~\cite{aspanformer}} & 43.4                           & 51.0                           & 54.0                           & 55.8                           & 51.8                           \\
\multicolumn{1}{l|}{Ours}        & \textbf{48.5} & \textbf{58.5} & \textbf{61.1} & \textbf{61.1} & \textbf{57.2} \\ \hline
\multicolumn{6}{c}{Matching coverage (\%)}                                                                                                                                                                      \\ \hline
\multicolumn{1}{l|}{SP\cite{superpoint}+SuperGlue~\cite{superglue}  }   & 13.1                           & 20.2                           & 23.7                           & 22.0                           & 16.1                           \\
\multicolumn{1}{l|}{LoFTR~\cite{loftr} }       & 21.4                           & 47.2                           & 65.8                           & 70.9                           & 35.8                           \\
\multicolumn{1}{l|}{ASpanFormer~\cite{aspanformer}} & 24.7                           & 50.0                           & 68.4                           & 83.7                           & 62.9                           \\
\multicolumn{1}{l|}{Ours}        & \textbf{77.3} & \textbf{89.5} & \textbf{90.8} & \textbf{88.4} & \textbf{81.1} \\ \hline
\multicolumn{6}{c}{Matching precision (\%)}                                                                                                                                                           \\ \hline
\multicolumn{1}{l|}{SP\cite{superpoint}+SuperGlue~\cite{superglue}  }   & 57.9                           & 67.9                           & 67.4                           & 63.0                           & 59.3                           \\
\multicolumn{1}{l|}{LoFTR~\cite{loftr}}       & 74.5                           & 77.5                           & 75.7                           & 65.6                           & 55.5                           \\
\multicolumn{1}{l|}{ASpanFormer~\cite{aspanformer}} & 77.7                           & 78.1                           & 76.6                           & 69.3                           & 63.1                           \\
\multicolumn{1}{l|}{Ours}        & \textbf{85.9} & \textbf{81.8} & \textbf{78.4} & \textbf{70.0} & \textbf{64.3} \\ \bottomrule
\end{tabular}
\vspace{-0.3cm}
\caption{Evaluation on extreme-scale dataset. Compared with other methods, PATS presents strong robustness against scale variations.}
\label{tab:megadepth}
\vspace{-0.5cm}

\end{table}

\subsection{Relative Pose Estimation}
\label{sec:pose_estimation}
We use ScanNet~\cite{scannet} and YFCC100M~\cite{yfcc100m} to evaluate the effectiveness of our method for relative pose estimation in both indoor and outdoor scenes, respectively.
Then to fully evaluate the performance under scale changes, we create an extreme-scale dataset by artificially scaling the images from MegaDepth~\cite{megadepth} dataset.

\noindent\textbf{Experimental setup.}
Following~\cite{superglue,loftr}, we train the indoor model of PATS on the ScanNet dataset, while training the outdoor model on MegaDepth.
We report the pose accuracy in terms of AUC metric at multiple thresholds ($5^\circ$,$10^\circ$,$20^\circ$).
For the evaluation on the extreme-scale dataset, we show two additional metrics to analyze factors that contribute to accurate pose estimation, including matching precision~\cite{superglue} and matching coverage~\cite{colmap}.
While the former metric measures the precision, the latter one describes how well the matched feature is distributed uniformly in the image.
% More details are provided in the supplementary material.

\noindent\textbf{Dataset.}
ScanNet is composed of 1613 indoor sequences.
We follow the same training and testing split used by \cite{superglue} , where 1.5K image pairs are used to evaluate.
YFCC100M~\cite{yfcc100m} contains a total of 100 million media objects.
We evaluate on a subset of YFCC100M, which consists of 4 selected image collections of popular landmarks following \cite{superglue,loftr}.
MegaDepth consists of one million Internet images of 196 outdoor scenes.
We sample 1000 image pairs and manually scale the second image to five different resolutions from 320 to 1600 along the longer side, which make up our extreme-scale dataset.

\begin{table}[!t]
% \vspace{-0.3cm}

% \tabcolsep 5pt
\footnotesize
% \normalsize

\centering
\begin{tabular}{lcc}
\toprule
\multirow{2}{*}{Method} &  DUC1                        & DUC2                       \\ \cline{2-3} 
                               & \multicolumn{2}{c}{(0.25m, 10\textdegree)\ /\ (0.5m, 10\textdegree)\ /\ (1m, 10\textdegree)} \\ \hline
SP~\cite{superpoint}+SuperGlue~\cite{superglue}                       & 49.0/68.7/80.8              & 53.4/\underline{77.1}/82.4             \\
LoFTR~\cite{loftr}                          & 47.5/\underline{72.2}/\underline{84.0}              & 54.2/74.8/\textbf{85.5}             \\
ASpanFormer~\cite{aspanformer}                    & \underline{51.5}/\textbf{73.7}/\textbf{86.0}              & \underline{55.0}/74.0/81.7             \\
Ours                       & \textbf{55.6}/71.2/81.0     & \textbf{58.8}/\textbf{80.9}/\textbf{85.5}    \\ \bottomrule
\end{tabular}
\vspace{-0.3cm}
\caption{Visual Localization on the InLoc benchmark~\cite{inloc}.}
\vspace{-0.4cm}
\label{tab:inloc}
\end{table}

\begin{table}[!t]
% \vspace{-0.5cm}
\footnotesize
\centering
\begin{tabular}{lcc}
\toprule
\multirow{2}{*}{Method} & \multicolumn{1}{c}{Day} & Night                                       \\ \cline{2-3} 
                                    & \multicolumn{2}{c}{(0.25m, 2\textdegree)\ / \ (0.5m, 5\textdegree)\ /\ (1m, 10\textdegree)}                \\ \hline
\multicolumn{3}{l}{Visual Localization of Aachen v1.0}                                                      \\ \hline
SP~\cite{superpoint}+SuperGlue~\cite{superglue}                           & ---                     & 79.6/\underline{90.8}/\textbf{100}                              \\
PDC-Net+~\cite{pdcnet++}                            & ---                     & 79.6/\underline{90.8}/\textbf{100}                               \\
PoSFeat\cite{posfeat}                             & ---                     & \underline{81.6}/\underline{90.8}/\textbf{100}                              \\
Ours                                & ---                     & \textbf{85.7/94.9/100}                      \\ \hline
\multicolumn{3}{l}{Visual Localization of Aachen v1.1}                                          \\ \hline
SP~\cite{superpoint}+SuperGlue~\cite{superglue}                           & \textbf{89.8/96.1/99.4} & \multicolumn{1}{c}{77/90.6/\textbf{100.0}}           \\
LoFTR~\cite{loftr}                               & 88.7/95.6/99.0          & \multicolumn{1}{c}{\textbf{78.5}/90.6/99.0} \\
ASpanFormer~\cite{aspanformer}                         & 89.4/95.6/99.0          & \multicolumn{1}{c}{\underline{77.5}/\underline{91.6}/\underline{99.5}}          \\
Ours                                & \underline{89.6/95.8/99.3}          & \multicolumn{1}{c}{73.8/\textbf{92.1}/\underline{99.5}}        \\ \bottomrule
\end{tabular}
\vspace{-0.3cm}
\caption{Visual Localization on the Aachen day-night benchmark~\cite{aachen}.}

\label{tab:aachen}
\vspace{-0.5cm}
\end{table}
\begin{table}[!t]
% \vspace{-0.5cm}

\tabcolsep 5pt
\footnotesize
% \normalsize

\centering
\begin{tabular}{cl|cccc}
\toprule

Training &\multirow{2}{*}{Method}          & \multicolumn{2}{c}{KITTI-2012} & \multicolumn{2}{c}{KITTI-2015} \\ \cline{3-6}
              Data &  & APAE       & Fl-all       & APAE        & Fl-all      \\ \hline
% \multirow{5}{*}{C+T}
\multirow{5}{*}{C + T} &PWC-Net~\cite{pwcnet}          & 4.14       & 20.28             & 10.35       & 33.67            \\
&GLU-Net~\cite{glunet}          & 3.34       & 18.93             & 9.79        & 37.52            \\
&RAFT~\cite{raft}             & ---        & ---               & 5.04        & 17.8             \\
&GLU+GOCor~\cite{gocor}    & 2.68       & 15.43             & 6.68        & 27.52            \\
&FlowFormer~\cite{flowformer}       & ---        & ---               & 4.09        & 14.72            \\ \hline
% \multirow{4}{*}{C+T}
\multirow{5}{*}{M} & PDC-Net+~\cite{pdcnet++}         & 1.76       & 6.6               & 4.53        & 12.62            \\
&COTR + Intp.~\cite{cotr}   & 1.47       & 8.79              & 3.65        & 13.65            \\
&ECO-TR + Intp.~\cite{eco} & 1.46       & 6.64              & \textbf{3.16}        & 12.10             \\
&Ours + Intp.   & \textbf{1.17}       & \textbf{4.04}              & 3.39        & \textbf{9.68}             \\
\bottomrule
\end{tabular}
\vspace{-0.3cm}
\caption{Optical flow estimation on the KITTI~\cite{kitti} benchmark. C + T indicates FlyingChairs and FlyingThings, and M indicates Megadepth.}
\label{tab:kitti}
\vspace{-0.4cm}

\end{table}

\begin{figure*}
\vspace{-0.8cm}
\begin{center}
\includegraphics[width=0.85\textwidth]{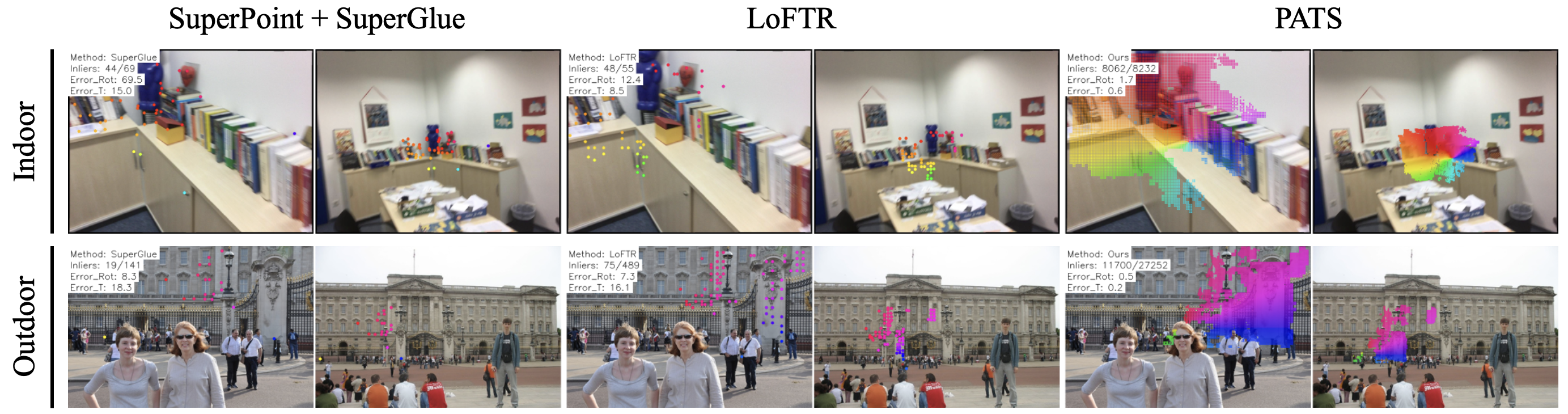}
\vspace{-0.5cm}

\end{center}
\caption{\textbf{Qualitative Comparison of Feature Matching.}
The matched features are visualized as the same color. We have filtered incorrect matches. PATS shows superior performance on both accuracy and coverage.
}
\vspace{-0.5cm}
\label{fig:results}
\end{figure*}

\noindent\textbf{Results.}
We compare PATS with both detector-based and detector-free methods. The detector-based methods have SuperPoint(SP)~\cite{superpoint} as feature extractor.
According to the results shown in Table~\ref{tab:yfcc} and Table~\ref{tab:scannet}, 
our method achieves state-of-the-art performance in both indoor and outdoor scenarios.
Next, we evaluate our approach against three representative methods using an extreme-scale dataset.
As presented in Table~\ref{tab:megadepth}, LoFTR exhibits a significant decrease in performance under extreme scale change, while our method presents strong robustness (e.g., LoFTR 22.2 vs. our 57.2 under 1600 resolution in terms of AUC)
ASpanFormer is an improved version of LoFTR, which still falls far short of our PATS at all resolutions.
The evaluation in terms of matching accuracy and matching coverage demonstrates that our PATS obtains matches that are more accurate and uniformly distributed in the image.
All of them contribute to the more accurate pose estimation.
We show the qualitative result in Fig.~\ref{fig:results}.
% and provide more results in the supplementary materials.

\subsection{Visual Localization}
Visual localization aims to recover 6-DoF poses of query images concerning a 3D model.
We use the InLoc~\cite{inloc} and Aachen Day-Night~\cite{aachen} datasets to validate the visual localization in indoor and outdoor scenes, respectively.

\noindent\textbf{Experimental setup.}
Following \cite{superglue,loftr,aspanformer}, we use the localization toolbox HLoc~\cite{hloc} with the matches computed by PATS
and evaluate on 
the LTVL benchmark~\cite{longterm}.
The performance is measured by the percentage of queries localized within multiple error thresholds.

\noindent\textbf{Dataset.}
The Aachen day-night v1.0 provides 4328 database images and 922 query images,
and provides additional 2359 database images
and 93 query images in v1.1.
A great challenge is posed in matching texture-less or repetitive patterns with a huge difference in perspective..
Inloc contains 9972 RGB-D images geometrically registered to the floor maps.
Great challenges are posed in identifying correspondences from images, in particular night-time scenes, which occur in extremely large illumination changes.

\begin{table}[!t]
\footnotesize
\centering
\begin{tabular}{l|cccc}
\toprule
Ablation study                            & AUC@5\textdegree & Coverage & Precision \\ \hline
L=2 w/o area regression           & 50.7  & 49.8          & 68.4         \\
L=2 w/o transportation & 40.1  & 52.8          & 66.7        \\
L=2 w/o concentration loss           & 41.6  & 32.1          &  75.9              \\
L=2 w/o outlier loss           & 36.3  & \textbf{98.4}          & 62.8              \\
L=2 w/o spliting M           & 50.9  & 92.4          & 72.1          \\
L=1                         &  0.7 &    5.8     & 55.5 \\
L=2                         & 51.9  & 92.5          & 72 \\
\textbf{L=3 (full)}                          &  \textbf{61.1}&   92.5& \textbf{78.4}  
\\ \bottomrule
\end{tabular}
\vspace{-0.2cm}
\caption{Abaltion studies on the Megadepth~\cite{megadepth} dataset.
% We evaluate our method on Megadepth dataset with each design or network component turned off. Overall, our full model achieves the best performance, which indicates the positive contribution from all design choices.
}
\label{tab:ablation}
\vspace{-0.4cm}

\end{table}

\noindent\textbf{Results.}
In Table~\ref{tab:inloc}, our method achieves the overall best results on InLoc dataset and outperforms other methods by a large margin in terms of percentage within the minimum error threshold (51.5 vs. our 55.6 in DUC1 and 55.0 vs. our 58.8 in DUC2).
We report the results on Aachen day-nigh in Table~\ref{tab:aachen} where our method achieves state-of-the-art performance in the challenging night track, especially on the v1.0 (81.6 vs. our 85.7 and 90.8 vs. our 94.9 in the first two metrics).
On the day track of v1.1, our method outperforms all other methods but is slightly inferior to SuperGlue.
% The details are provided in the supplementary.

\subsection{Optical Flow Estimation}
We also evaluate our model on optical flow, which estimates per-pixel correspondences between an image pair.

\noindent\textbf{Experimental setup.}
Our evaluation metrics include the Average End-point Error (AEPE) and Fl-all.
Fl-all denotes the proportion of pixels for which the flow error exceeds either 3 pixels or 5\% of the length of the ground truth flows.
Following~\cite{cotr,eco}, we interpolate our model's output to obtain per-pixel correspondences (``+intp.'')  and evaluate the accuracy only on those non-occluded pixels. %thus the results are not directly comparable to those optical flow methods such as RAFT~\cite{raft} and FlowFormer~\cite{flowformer}. 
Different from \cite{cotr,eco} that further filter out the points which do not satisfy the cycle-consistency constraint, we do not conduct this operation but still achieve better accuracy.
% Apart from that, we don't filter out points that do not satisfy the cycle-consistency constraint.

\noindent\textbf{Dataset.}
KITTI datasets are collected in urban traffic scenes.
KITTI-2012 dataset contains simpler scenarios, while KITTI-2015 dataset has challenging and dynamic scenarios.
Following \cite{glunet,cotr}, we evaluate on the KITTI training sets without finetuning.

\noindent\textbf{Results.}
We divide the methods into two categories for comparison. One is trained on the FlyingChairs~\cite{flyingchair} and FlyingThings~\cite{flyingthings}, specially designed for optical flow, and the other is trained on MegaDepth.
As shown in Table~\ref{tab:kitti}, our method obtains overall state-of-the-art performance.
In terms of Fl-all metric, our method outperforms second-best method (ECO-TR) by a large margin in both KITTI-2012(+39\%) and KITTI-2015(+20\%).

\subsection{Ablation Studies}
To understand the effect of each design on the overall structure, we conduct the ablation study on MegaDepth dataset by disabling each design individually.
In Table~\ref{tab:ablation}, we show the quantitative results.
For the sake of the training time, we conduct most of the ablation studies only in the second hierarchy (denoted as $L=2$), fixing the first hierarchy and removing the third one.

\noindent\textbf{Removing the Area Regression.} 
We re-train a model by removing the area regression (denoted as ``w/o area regression''). The performance drop indicates that modeling the area relationship positively contribute to the model.

\noindent\textbf{Removing Transportation.}
We also try removing the entire module of patch area transportation (denoted as ``w/o transportation''). In this case when computing the corresponding position of each source patch, we compute the expectation based on the feature similarity instead of the transported area (Eq.~\ref{eq:interpolate}), which is the similar to LoFTR.
The performance degradation shows the importance of the proposed area transportation.

\noindent\textbf{Impact of Each Loss Term.}
We study the impact of concentration loss and outlier patch loss by disabling them respectively and we also study the impact of dividing inlier and outlier patch (denoted as ``w/o spliting M''). Note that the regression supervision in the inlier patch loss is the basic of our loss function so we do not remove them in all the ablation experiments.
Overall, removing any design result in a decrease in the accuracy of the pose estimation.

\noindent\textbf{Number of Hierarchies.}
Besides, we show the impact of increasing the number of hierarchies (denoted as ``L=1,2,3'', respectively).
It can be seen that the performance keeps improving as the number of hierarchies increases.

\section{Conclusion}
\label{sec:conclusion}

In this paper, we propose Patch Area Transportation with Subidivion~(PATS) for local feature matching.
PATS learns to find the many-to-many relationship and scale differences through the proposed patch area transportation. Accompanied by the proposed patch subdivision algorithm, PATS enables extracting high-quality semi-dense matches even under severe scale variations.
% By splitting images into patches, PATS estimate their scale differences and aligns their contents according to the scales so that the appearance distortion derived from scale change is alleviated and the resultant feature matches are more accurate.
% Moreover, PATS also allows multiple features in one image to match features in very close proximity to the other image, which largely increases the number of matches. These are enabled by the proposed novel patch area transportation accompanied by the patch subdivision algorithm.
% PATS achieves significantly better correspondence quality, in terms both of accuracy and coverage, and benefits downstream tasks, including relative pose estimation, visual localization, and optical flow.
Multiple datasets demonstrate that PATS delivers superior performance in relative pose estimation, visual localization, and optical flow estimation, surpassing state-of-the-art results.
One drawback of PATS is its inability to deliver real-time performance. However, it is promising to enhance our model and expanding the system to accommodate real-time applications such as SLAM.

\noindent\textbf{Acknowledgements.}
This work was partially supported by NSF of China (No. 61932003).

\normalem

{\small
\bibliographystyle{ieee_fullname}
\bibliography{egbib}
}

\end{document}